\documentclass[letterpaper, 10 pt, conference]{ieeeconf}
\IEEEoverridecommandlockouts

\usepackage[english]{babel}
\usepackage{afterpage}
\usepackage{placeins}
\usepackage{amsmath, amssymb, amsfonts}
\usepackage{graphicx}
\usepackage{lipsum}
\usepackage{xcolor}
\usepackage{soul}
\usepackage{makecell}
\usepackage{booktabs}
\usepackage{array}
\usepackage{cite}
\usepackage{subcaption}

\usepackage[colorlinks=true, allcolors=blue]{hyperref}

\definecolor{customPurple}{RGB}{112, 48, 160}
\definecolor{mutedblue}{RGB}{219,234,253}
\definecolor{mutedgreen}{RGB}{216,235,205}
\definecolor{mutedorange}{RGB}{251,229,214}
\usepackage{amssymb}
\usepackage{pifont}
\title{\LARGE \bf
Language as Cost: Proactive Hazard Mapping using VLM\\
for Robot Navigation
}

\author{
    Mintaek Oh, Chan Kim, Seung-Woo Seo and Seong-Woo Kim%
    \thanks{All authors are with the Seoul National University, Seoul, Republic of Korea.
    {\tt\small \{mintaek,chan\_kim,sseo,snwoo\}@snu.ac.kr}}%
}

\date{}

\begin{document}
\maketitle

\begin{abstract}
Robots operating in human-centric or hazardous environments must proactively anticipate and mitigate dangers beyond basic obstacle detection. Traditional navigation systems often depend on static maps, which struggle to account for dynamic risks, such as a person emerging from a suddenly opening door. As a result, these systems tend to be reactive rather than anticipatory when handling dynamic hazards.
Recent advancements in pre-trained large language models and vision-language models (VLMs) create new opportunities for proactive hazard avoidance. In this work, we propose a zero-shot language-as-cost mapping framework that leverages VLMs to interpret visual scenes, assess potential dynamic risks, and assign risk-aware navigation costs preemptively, enabling robots to anticipate hazards before they materialize. 
By integrating this language-based cost map with a geometric obstacle map, the robot not only identifies existing obstacles but also anticipates and proactively plans around potential hazards arising from environmental dynamics. Experiments in simulated and diverse dynamic environments demonstrate that the proposed method significantly improves navigation success rates and reduces hazard encounters, compared to reactive baseline planners. Code and supplementary materials are available at \href{https://github.com/Taekmino/LaC}{https://github.com/Taekmino/LaC}.

\end{abstract}

\section{Introduction}

\afterpage{%
    \begin{figure}[h]
        \centering
        \includegraphics[width=\columnwidth]{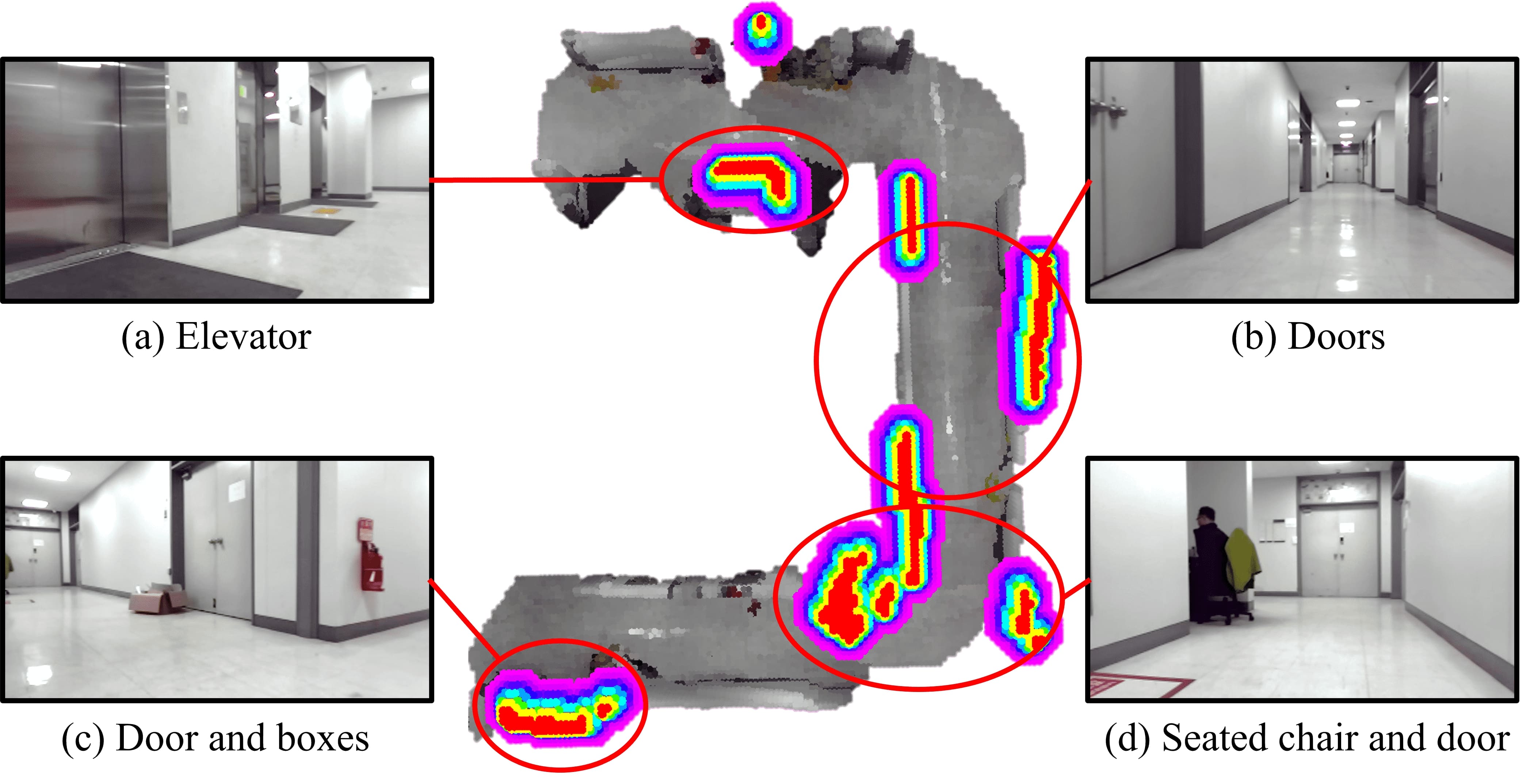}
        \caption{\label{fig:bev_map}Real-time generated Gaussian map using the proposed methodology. Potential risks are inferred from each image without direct experience, and an anxiety score is pre-computed using a VLM to update the cost of the map. The anxiety score determines how much the robot should avoid certain areas during navigation. These inferred hazards are then integrated into the cost map, allowing the robot to proactively adjust its navigation strategy based on potential risks.}
    \end{figure}
}

Mobile robots are increasingly deployed in everyday environments, such as homes, hospitals, warehouses, and disaster sites, where safety and context-aware navigation are critical. Most robots navigate these spaces by constructing static obstacle maps and avoiding obstacles to ensure collision-free movement. However, static obstacle maps are often insufficient in environments where unrepresented hazards may emerge and impede the robot's navigation. For example, a static obstacle map cannot enable a robot to anticipate a door from which a person may suddenly step out, posing a potential risk. Although robots can avoid visible dynamic obstacles by integrating object detection, they may still overlook less apparent yet potentially severe hazards that require a deeper understanding of the environment \cite{jeon2025non}.

Recent studies on hazard detection have proposed various approaches to overcome this limitation. Risk-aware navigation algorithms, such as RAMP~\cite{sharma2023ramp}, extended traditional 2.5D occupancy grids by incorporating variable-horizon planning, allowing robots to navigate unknown or uncertain regions with an adaptive balance between speed and caution. Other approaches explicitly incorporate semantic context. Ashour \textit{et al.} \cite{ashour2022semantic} proposed a hazard identification and risk assessment system that fuses deep learning-based semantic labels into a 3D map, thereby enabling more nuanced risk scoring than simple geometry-based cost maps. Likewise, efforts in visual anomaly detection for mobile robots \cite{6386031, mantegazza2022outlier, salimpour2022selfcalibratinganomalychangedetection} treat any out-of-distribution observation as a potential hazard. By learning what normal environments look like, the robot can identify unusual sights, such as unexpected obstacles or changes, as hazards to investigate or avoid. Although anomaly detection is effective in identifying unforeseen hazards, it lacks semantic understanding, such as recognizing the nature of a hazard. As a result, it struggles to assign varying risk levels to different hazards.

In parallel, robotics researchers have increasingly leveraged vision-language models (VLMs) and large language models (LLMs) as foundation models for hazard detection and failure reasoning. These models, pre-trained on vast visual and textual data, provide robots with robust perception and reasoning capabilities, enabling them to understand complex environments and instructions. Studies such as \cite{eskandari2025llm, duan2024aha} demonstrate how natural language descriptions can enhance hazard annotation in virtual environments and improve failure reasoning in robotic manipulation, offering valuable insights for both human operators and autonomous systems. Specifically, Eskandari \textit{et al.} \cite{eskandari2025llm} presented a system where a state-of-the-art VLM automatically annotates hazards in a virtual environment, enhancing situational awareness in high-risk scenarios. Meanwhile, Duan \textit{et al.} \cite{duan2024aha} proposed a vision-language model for detecting and reasoning about failures in robotic manipulation, providing detailed explanations that improve robotic system performance.

In this context, we explore how textual hazard cues can be seamlessly integrated into the cost function, enabling robots to proactively avoid potential risks. By prompting VLMs with hazard descriptions such as wet floor or no entry, the robot can translate these cues into a semantic costmap that guiding its navigation pipeline to circumvent risky areas rather than merely reacting to known obstacles. However, existing approaches often confine hazard handling to detection alone or require explicit user commands to reflect identified hazards on the map, thus missing the opportunity to proactively incorporate potential risks into the robot’s path planning.

To truly enhance robotic hazard detection, it is essential to move beyond mere perception and incorporate proactive risk assessment in navigation. Humans often avoid hazards through logical inference and prior knowledge without requiring direct experience for every new situation. Even in unseen environments, individuals can anticipate threats by combining observational cues with a reasoning process that infers hidden dangers, such as a door that might open suddenly, a passing car in an unseen alley, or an icy patch on a road. Although past experiences certainly inform these inferences, not every hazard need to be learned through trial and error. Hartley and Phelps \cite{hartley2012anxiety} suggested that anxiety plays a key role in this process. Individuals with higher anxiety levels generally adopt more cautious decision-making in uncertain conditions, tending to avoid potential risks proactively. Romain \cite{romain2024decisions} further suggested that state-induced anxiety may enhance decision-making performance by increasing attentional focus on potential risks, though its effects are context-dependent. These findings underscore that humans’ ability to foresee and evade threats arises not merely from direct encounters but also from an intrinsic psychological mechanism that evaluates possible risks in advance.

Inspired by this capacity for human-like inference, we propose a framework that proactively encodes potential dangers into a cost map without requiring prior experience of hazards. By leveraging VLMs, the system identifies both objects and areas that may pose risks, such as corners, blind spots, and slippery surfaces, assigning elevated costs through language-driven inference and visual-semantic understanding. To accurately localize these hazards in the cost map, we employ a zero-shot semantic segmentation model, Grounded Edge SAM \cite{zhou2024edgesampromptintheloopdistillationondevice}, which detects risky objects at the pixel level and integrates them with depth information. Furthermore, to accommodate environmental changes, we incrementally update the cost map during navigation to incorporate newly detected threats. Consequently, the robot’s planning and control modules operate with greater caution, mirroring humans’ ability to anticipate and avoid hazardous conditions before they arise. The proposed framework follows a Bayesian spatial cognition approach, leveraging the historically accumulated linguistic and cultural knowledge of potential risk situations embedded in VLMs as a priori abstract spatial cost function, while continuously updating it through experience as a posterior refinement.

The contributions are summarized as follows:  
\begin{itemize} 

\item We introduce a zero-shot language-as-cost mapping framework that utilizes VLMs to analyze visual scenes, evaluate potential dynamic risks, and proactively assign risk-aware navigation costs, enabling robots to anticipate and proactively respond to hazards.
\item By continuously updating the cost map during navigation to incorporate newly detected potential threats, we demonstrate that the proposed method ensures adaptive and safe navigation.
\end{itemize}

\renewcommand{\arraystretch}{2.0}
\begin{table}[t]
\centering
\caption{Comparison of Existing and Proposed Methods.}
\label{table:comparison}
\resizebox{1.0\linewidth}{!}{
\begin{tabular}{c|ccccc}
Methods &      
    \makecell{\textbf{Hazard} \\ \textbf{awareness}} & 
    \makecell{\textbf{Semantic} \\ \textbf{understanding}} & 
    \makecell{\textbf{Zero-shot} \\ \textbf{generalization}} &
    \makecell{\textbf{Open-set} \\ \textbf{categories}} &
    \makecell{\textbf{Dynamic} \\ \textbf{potential hazard}} 
    \\ \hline \hline
    
    \makecell{\cite{pao2025pfbcp}} & \checkmark & \checkmark & & &  \\ \hline    
    \makecell{\cite{sharma2023ramp, qi2020learning}} & \checkmark & & & & \ensuremath{\triangle}   \\ \hline
    \makecell{\cite{sun2023risk}} & \checkmark & & & & \checkmark \\ \hline
    \makecell{\cite{cai2023probabilistic, aegidius2025watch}} & \checkmark & \ensuremath{\triangle} & \ensuremath{\triangle} & &  \\ \hline
    \makecell{\cite{kim2024e2map}} & \checkmark & \checkmark & & \checkmark & \ensuremath{\triangle} \\ \hline
    \makecell{\cite{chen2024affordances}} & \checkmark & \checkmark & \checkmark & \ensuremath{\triangle} &  \\ \hline
    \makecell{\cite{bolte2023usa}} & \checkmark & \checkmark & \checkmark & \checkmark & \\ \hline

    \specialrule{0.1em}{0em}{0em}
    \textbf{Ours} & \checkmark & \checkmark & \checkmark & \checkmark & \checkmark \\ \hline
\end{tabular}
}
\vspace{-1.0em}
\end{table}

\section{Related Works}

\subsection{Risk-aware Navigation}

Recent advancements in risk-aware navigation research have focused on exploring probabilistic modeling, self-supervised learning, reinforcement learning and large language models. 

Probabilistic Modeling-based approaches probabilistically modeled hazardous elements in the environment and performed path planning that reflected environmental uncertainty. 
Pao \textit{et al.} \cite{pao2025pfbcp} improved real-time risk assessment by modeling scene affordance through potential fields, using BEV semantic segmentation to enhance spatial accuracy, temporal consistency, and computational efficiency. However, their approach did not incorporate semantic scene understanding, resulting in a lack of generalization ability for dynamic and semantically complex environments, as well as limitations in zero-shot generalization and open-set category handling. 
Cai \textit{et al.} \cite{cai2023probabilistic} developed a probabilistic traction model for off-road navigation, utilizing self-supervised learning to estimate terrain slip and uncertainty for risk-aware path planning. Nevertheless, their method was limited in semantic understanding and zero-shot generalization capabilities, and did not address open-set categories or dynamic potential hazards.

Self-Supervised Learning-based approaches focused on improving real-time navigational adaptation by leveraging unlabeled environmental data. 
STEPP \cite{aegidius2025watch} employed self-supervised learning to estimate terrain traversability and adapt navigation in real time, using vision-based features to detect anomalies. However, it was primarily reactive and relied on prior training data, limiting its ability for full zero-shot generalization. 
Similarly, Cai \textit{et al.} \cite{cai2023probabilistic} leveraged self-supervised learning alongside a probabilistic traction model to predict terrain slip. Both approaches did not perform open-set category detection and failed to consider dynamic potential hazards.

Reinforcement Learning-based approaches optimized the robot’s policy to learn traversable regions or proactively avoid risks based on reward signals. 
Qi et al. \cite{qi2020learning} proposed an RL-based method for learning navigable regions using spatial affordance maps. However, their approach lacked semantic object classification and real-time hazard assessment, and was limited in zero-shot generalization and open-set category handling. It only partially addressed dynamic potential hazards.
Sun \textit{et al.} \cite{sun2023risk} proposed a risk-aware RL framework that dynamically adjusted navigation based on pedestrian collision risk, achieving high success rates in dense crowds. Despite this, their method demonstrated limited generalization performance in previously unseen environments due to the absence of semantic scene understanding, open-set category handling, and zero-shot generalization.

Semantic approaches in navigation leverage CLIP-based models and open-vocabulary representations to provide robots with flexible semantic understanding and language-based goal specification.
Bolte \textit{et al.} \cite{bolte2023usa} proposed a unified 3D implicit map integrating semantic information and spatial affordance through CLIP embeddings and signed distance fields, enabling efficient open-vocabulary navigation and motion planning. However, their method did not address hazard awareness or dynamic potential hazards, as it is designed for static environments without modules for real-time hazard detection or risk forecasting.

In contrast, recent approaches have begun harnessing LLMs and VLMs to augment risk-aware navigation with enhanced semantic understanding. 
Chen \textit{et al.} \cite{chen2024affordances} leverage LLMs to integrate affordance-based navigation with waypoint selection and motion control in continuous environments, focusing on collision-free path planning while lacking proactive hazard anticipation. 
E2Map \cite{kim2024e2map} leverages LLMs and VLMs to encode unexpected collisions or events as emotional responses, which are stored in a spatial map and used for future path planning. While this approach improves adaptability by integrating past experiences, it updates only when events occur, making it risky as hazards must be encountered first. The distinctive aspects of our approach compared to previous works are detailed in Table \ref{table:comparison}.

\subsection{Grounded SAM}

SAM \cite{kirillov2023segment} is designed to rapidly and broadly segment regions within an image using minimal prompts, for instance, a single point or a bounding box. By performing generic segmentation across the entire image, SAM can quickly delineate various objects without being constrained by specific classes. This flexibility proves highly advantageous in real-time robot navigation, where users may supply arbitrary prompts and demand swift object recognition. However, SAM does not provide a built-in mechanism for text-based segmentation prompts.

To overcome this limitation, Grounded SAM \cite{ren2024grounded} integrates the open-set detection capabilities of Grounding DINO \cite{liu2024grounding} with SAM’s prompt-based segmentation. Grounding DINO identifies bounding boxes for arbitrary objects specified via text prompts, while SAM refines these regions at the pixel level. This combined approach effectively enhances open-world visual perception by recognizing objects beyond predefined labels and enabling seamless integration with other specialized visual models. Grounded SAM has demonstrated strong performance in segmentation benchmarks, highlighting its effectiveness in diverse environments.

Our framework leverages VLMs to dynamically identify potential risk objects based on the current context. Since these objects may not correspond to predefined classes, conventional semantic segmentation models constrained to fixed labels are inadequate. Instead, zero-shot segmentation methods, particularly Grounded Edge SAM \cite{zhou2024edgesampromptintheloopdistillationondevice} for efficient inference, enable the recognition of novel, language-driven categories. By integrating the VLM’s adaptive object detection with a zero-shot segmentation layer, our approach achieves both high flexibility and speed, ensuring rapid identification of unexpected or newly introduced hazards.

\section{Method}

\begin{figure*}[ht]
\centering
\includegraphics[width=\textwidth]{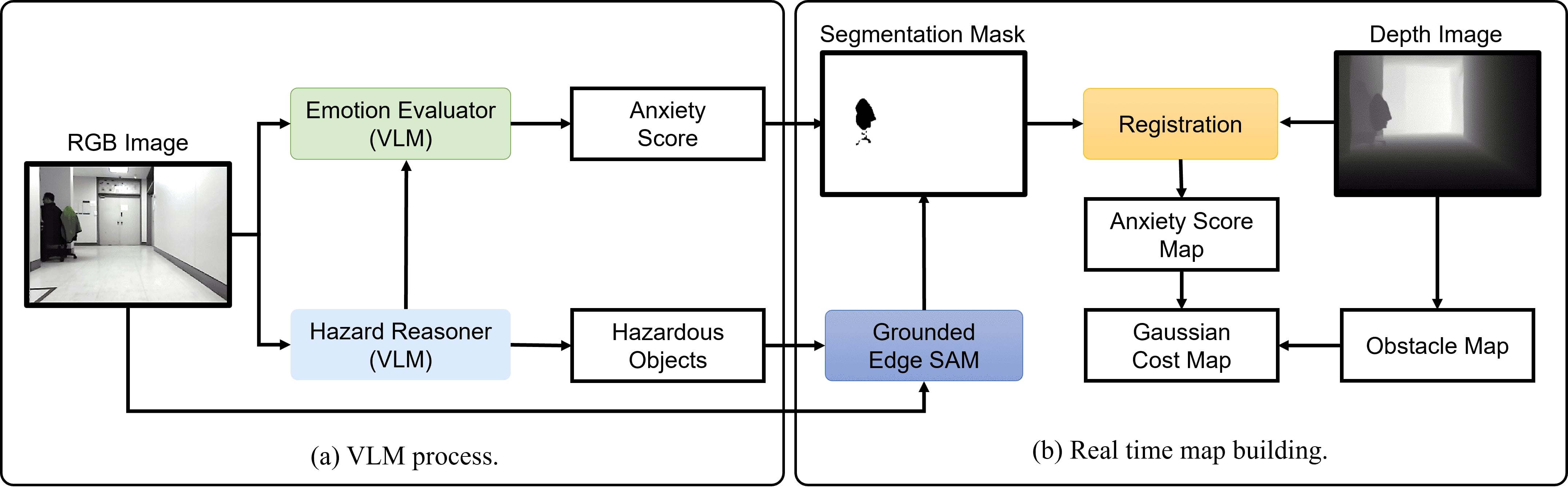}
\caption{\label{fig:arch}System architecture. (a) The hazard reasoner describes the scene from the robot's viewpoint, identifying potentially hazardous situations and objects. The emotion evaluator then computes the anxiety score of hazardous objects based on the hazard reasoner’s output and the robot’s viewpoint image. (b) A real-time Gaussian map is generated to reflect potential hazards. The hazardous objects identified by the hazard reasoner are segmented using a zero-shot segmentation model to extract masks, which are combined with the anxiety scores to create an anxiety score map. This map is then integrated with the obstacle map to generate the Gaussian cost map.}
\vspace{-1.0em}
\end{figure*}

\begin{figure}[t]
\centering
\includegraphics[width=\columnwidth]{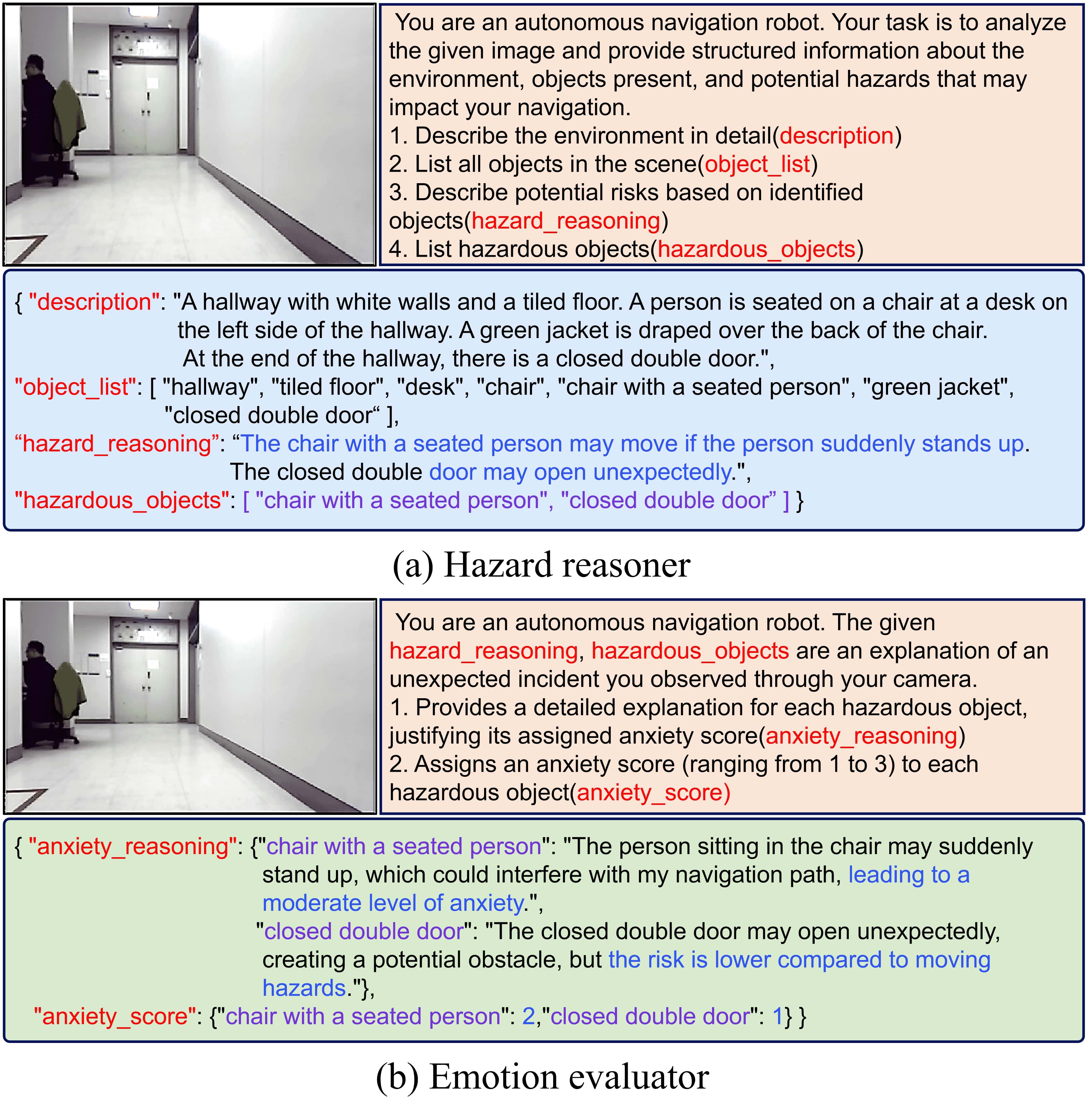}
\caption{\label{fig:VLM}Each top box is the VLM's input, and the bottom box is the output.  
The \fcolorbox{mutedorange}{mutedorange}{\makebox[1.5cm]{orange box}} represents the VLM's system prompt.  
The \fcolorbox{mutedblue}{mutedblue}{\makebox[1.3cm]{blue box}} is the output of (a) hazard reasoner,  
and the \fcolorbox{mutedgreen}{mutedgreen}{\makebox[1.3cm]{green box}} is the output of (b) emotion evaluator.  
All outputs are in JSON format.}
\label{fig:VLMs}
\vspace{-2.0em}
\end{figure}

Our method aims to proactively identify and avoid hazards during navigation by leveraging a VLM-based \textit{Hazard Reasoner} and \textit{Emotion Evaluator}, followed by zero-shot segmentation and Gaussian-based cost propagation. The system architecture is shown in Fig. \ref{fig:arch}.

First, we prompt the VLM with a hazard reasoning prompt to obtain a list of potential risks in the scene, which is then refined by an emotion evaluator that assigns an anxiety score to each identified hazard. The VLM process is repeated after each inference cycle, and each inference takes approximately four seconds. The input and output of each VLM are shown in Fig. \ref{fig:VLMs}.

Next, the hazards are localized in the robot’s view via a zero-shot segmentation model, generating masks that merge with depth data to form a preliminary anxiety score map. At this stage, the model's input consists of the real-time RGB image and the most recently identified hazardous objects. Simultaneously, an obstacle map is generated in real time using depth images.

Finally, each hazardous cell’s anxiety score is propagated across neighboring regions through a two-dimensional Gaussian function, and the resulting cost map is fused with an obstacle map using a max-fusion strategy. This yields a final navigational map \(M_{\mathrm{Gaussian}}\), with values in the range [0,1], where higher values indicate more critical or impassable areas. The following subsections detail each component of the pipeline, from hazard reasoning to map integration.

\subsection{Hazard Situation Reasoning with VLMs}
In the first step, we utilize GPT-4o as the \textit{Hazard Reasoner} to identify potential threats and assess their severity, while GPT-4o-mini serves as the \textit{Emotion Evaluator} to quantify the level of risk associated with each hazard. By prompting the model with open-ended questions about possible dangers in the current view, we generate a structured hazard list that serves as input for downstream processing, as illustrated in Fig. \ref{fig:VLMs}(a). Additionally, the model assigns an anxiety score to each identified hazard, providing a measure of risk severity, as shown in Fig. \ref{fig:VLMs}(b). This VLM-based framework is inspired by multimodal Chain-of-Thought (CoT) \cite{zhang2024multimodalchainofthoughtreasoninglanguage}, which enhances interpretability and enables the system to refine its hazard assessments through step-by-step logical inference.

\subsubsection{Hazard Reasoner}
The hazard reasoner is a module that enables the robot to identify and evaluate potential hazards in its environment. By analyzing visual information, it allows the robot to recognize objects and contextual elements that may pose risks to navigation. These include dynamic hazards, such as a closed door behind which someone might suddenly appear, and static hazards, such as a wet floor that could cause instability. By structuring its reasoning process, the hazard reasoner ensures that the robot can logically interpret and respond to threats.
To achieve this, the robot captures an image \( I_t \) at time \( t \) and prompts the VLM with a hazard reasoning prompt \( p_{hd} \), which guides the model in identifying and reasoning about potential hazards. This process is formally defined as follows:
\[
J_t = F_{hd}(I_t,\, p_{hd}),
\]
where \( J_t \) is a structured JSON output that encapsulates multiple levels of reasoning about potential hazards in the scene. It consists of four key components: \textit{textual description}, \textit{object list}, \textit{hazard reasoning}, and \textit{hazardous objects}.
First, the \textit{textual description} provides a detailed textual representation of the environment in the image, offering contextual understanding. The \textit{object list} includes all detected objects within the scene, forming the foundation for risk assessment. Building upon this, the \textit{hazard reasoning} \(R_t\) component applies CoT reasoning to analyze potential hazards, linking objects and environmental factors to possible risks. Finally, the \textit{hazardous objects} \(L_t\) field contains a refined list of objects identified as significant threats to the robot’s navigation, derived from the preceding reasoning steps. This structured approach ensures interpretability in hazard detection, allowing the robot to make informed navigation decisions based on logical inference.

\subsubsection{Emotion Evaluator}
The emotion evaluator is a module that assesses the anxiety level associated with hazardous objects detected in an environment. It takes three inputs. The first input is \( R_t \), which represents \textit{hazard reasoning} from the hazard reasoner. The second input is \( L_t \), which is the list of detected \textit{hazardous objects}. The third input is \( I_t \),  which is the robot's viewpoint image at time \(t\). Additionally, the system prompt, denoted as \( p_{ee} \), guides the evaluation process. Based on these inputs, the emotion evaluator assigns an anxiety score to each detected hazardous object. The anxiety score is computed as follows:
\[
s_{\text{anxiety},t} = F_{ee}(R_{t},\ L_{t},\ I_t,\ p_{ee}),
\]
where \( s_{\text{anxiety},t} \) represents the anxiety score at time \( t \). By adopting a scoring scale and prompting the VLM to justify each score using a CoT approach, the emotion evaluator quantifies the potential severity of different hazards. This evaluation method is adapted from E2Map \cite{kim2024e2map}.

Each hazardous object is assigned an anxiety score on a scale of one and three. A higher anxiety score increases the cost around the object in the Gaussian cost map, ensuring that the robot avoids the area when planning its path. Furthermore, the anxiety score is used in the covariance calculation for the Gaussian map, shaping the cost distribution and influencing the robot’s navigation strategy to minimize exposure to high-risk areas.

\subsection{Language-based Segmentation Mask Extraction}

After identifying which objects or regions are potentially dangerous and assigning each an anxiety score, we create a language cost map to reflect these hazards in spatial coordinates.

We utilize Grounded Edge SAM \cite{zhou2024edgesampromptintheloopdistillationondevice}, a zero-shot segmentation model, to generate segmentation masks for the RGB image \( I_t \) based on the hazardous objects list \(L_t\). Specifically, \[
M_{t+k} = \text{Seg}(I_{t+k}, L_t),
\]
where \( M_{t+k} \) denotes the segmentation masks corresponding to each identified hazard. Each mask is labeled with its associated anxiety score \( s_{ \text{anxiety,t}} \), ensuring that the severity of potential risks is encoded in the map. This approach enables the localization of hazards in pixel space without requiring direct training for every possible object class. Since the inference speed of the VLM hazard reasoner is slower than that of the segmentation model, we introduce the notation \( (\cdot)_{t+k} \) to indicate the time index between the VLM’s output at time \(t\) and the subsequent segmentation inference before the next VLM's output update. This ensures that hazard-aware segmentation remains responsive, even when hazard reasoning operates at a lower frequency than visual processing.

\subsection{Gaussian Map Building}
\label{sec:map_integration}

\subsubsection{Map Representation}
We represent the environment as a top-down grid map 
\(
M \in \mathbb{R}^{\overline{H} \times \overline{W}},
\)
where \(\overline{H}\) and \(\overline{W}\) denote the number of rows and columns, respectively. Each cell corresponds to a physical location with resolution \(s\) meters per cell; thus, the total map size is \(s \,\overline{H} \times s \,\overline{W}\) meters. We refer to individual grid cells using the notation \(\mathbf{p}_i = (x_i,\, y_i)\), where \(x\) and \(y\) are the integer grid indices.  

\subsubsection{Obstacle Map}
In parallel with the VLM processes, the robot or system typically maintains an obstacle map, generated either from depth-based methods (e.g., depth images) or by constructing occupancy grid maps using a SLAM algorithm. This map indicates for each grid cell whether it is physically traversable or blocked by an obstacle. Formally, let
\[
M_{\mathrm{obs}}(\mathbf{p}_i) = 
\begin{cases}
1 & \text{if an obstacle is present at cell } \mathbf{p}_i,\\
0 & \text{otherwise}.
\end{cases}
\]
By assigning each obstacle cell a cost of \(1\) (the maximum in our normalized range of \([0,\,1]\)), we ensure that blocked areas are strictly avoided during path planning.

\subsubsection{Anxiety Score Map}
To incorporate language-based hazard information, we merge the segmentation masks with depth data from the robot’s sensors, thereby creating a localized 3D point cloud of detected hazards, where each point is associated with an anxiety score \( s_{\text{anxiety}} \). We then project these 3D points onto the 2D grid map, assigning each grid cell \( p_i \) a score \( a_i \) based on the projected points. To ensure consistency, the anxiety score map is generated using the same grid size as the previously constructed obstacle map. To remove noise and partial mismatches, we apply filtering before projection. This process yields a preliminary anxiety score map, where \( a_i \) is determined as a base cost proportional to \( s_{\text{anxiety}} \) and takes values from the set \( \{0, 1, 2, 3\} \).

\subsubsection{Gaussian Map Integration}
Beyond the binary hazard labeling, we propagate each cell’s anxiety score over neighboring regions by means of a two-dimensional Gaussian spread. This allows areas adjacent to a high-anxiety cell to inherit an elevated cost while still maintaining continuous variations.

\paragraph{Multivariate Gaussian Definition}
For each grid cell \(\mathbf{p}_i\) with a nonzero anxiety score \(a_i\), we define a multivariate Gaussian distribution
\(
\mathcal{N}\!\bigl(\mathbf{x}\,\big|\;\boldsymbol{\mu}_{\mathbf{p}},\,\Sigma_{\mathbf{p}}\bigr),
\)
where \(\mathbf{x}\) represents the continuous coordinates within the map, and
\renewcommand{\arraystretch}{1.0}
\(
\boldsymbol{\mu}_{\mathbf{p}} = 
\begin{bmatrix}
x & y
\end{bmatrix}
,\quad
\Sigma_{\mathbf{p}} = 
\begin{bmatrix}
\sigma_{x}^2 & 0; 0 & \sigma_{y}^2
\end{bmatrix}.
\)
We initialize \(\Sigma_{\mathbf{p}}\) as an identity matrix scaled by a coefficient \(\sigma_0^2\), and adjust \(\sigma_0\) based on factors such as the anxiety score. Higher scores yield larger \(\sigma\) values, causing the cost to spread more broadly.

\paragraph{Anxiety-Based Covariance Update}
After initialization, we further refine each diagonal element of the covariance matrix based on the associated \emph{anxiety score}. Inspired by the Weber--Fechner law~\cite{dehaene2003neural}, which posits that perceived intensity grows as a logarithmic function of the stimulus, we update the standard deviation for each axis \(k \in \{x_i, y_i\}\) as follows:
\[
\sigma_k^{\mathrm{new}}
=
\sigma_k 
\log\Bigl(
\tfrac{a_i}{T}
\Bigr),
\]
where \(a_i\) is the anxiety score, and \(T\) is a temperature parameter controlling the degree of spread. The function \((\cdot)^{\mathrm{new}}\) denotes the updated value. A higher \(a_i\) thus leads to a larger \(\sigma_k^{\mathrm{new}}\), increasing the spatial influence of that hazard.

\paragraph{Gaussian Cost Propagation.}
To embed the anxiety within the map, we compute each Gaussian’s contribution to a cell \(\mathbf{p}_j\) by evaluating:
\[
\mathrm{Cost}_i(\mathbf{p}_j) 
= 
\exp\!\Bigl(\!-\tfrac{1}{2}\,\tfrac{\|\mathbf{p}_j - \boldsymbol{\mu}_{\mathbf{p}_i}\|^{2}}{\sigma_i^{2}}\Bigr)\;\times\;\bigl(\tfrac{a_i}{3}\bigr),
\]
where \(\|\mathbf{x}_j - \boldsymbol{\mu}_{\mathbf{p}_i}\|\) is the Euclidean distance between the continuous center of cell \(\mathbf{p}_j\) and \(\boldsymbol{\mu}_{\mathbf{p}_i}\). We multiply by \(\frac{a_i}{3}\) to normalize the integer anxiety score to \((0,\,1]\). We then clamp cost to \([0,\,1]\).  

If multiple hazards overlap in cell \(\mathbf{p}_j\), we take the \emph{maximum} cost contribution:
\[
M_{\mathrm{cost}}(\mathbf{p}_j) 
= 
\max_{i\in \mathcal{H}}\;\mathrm{Cost}_{i}(\mathbf{p}_j),
\]
where \(\mathcal{H}\) is the set of all hazard cells with nonzero anxiety scores.

\paragraph{Map Integration}
We then merge the obstacle map \(M_{\mathrm{obs}}\) and the continuous cost map \(M_{\mathrm{cost}}\) into the final map \(M_{\mathrm{Gaussian}}\) by again applying a maximum operator:
\[
M_{\mathrm{Gaussian}}(\mathbf{p}_j) 
= 
\max\Bigl(M_{\mathrm{obs}}(\mathbf{p}_j),\, M_{\mathrm{cost}}(\mathbf{p}_j)\Bigr).
\]
Here, obstacles remain at cost \(1\), guaranteeing that they are strictly avoided, while other regions reflect the propagated cost values from the Gaussian spreads. By adopting this max-fusion strategy, we ensure that highly anxious or clearly blocked cells are assigned the highest priority in the navigation framework. The end result is a map with values in the normalized range \([0,\,1]\), in which \(1\) indicates an impassable region, and smaller values represent correspondingly lower risk levels.

\section{Experiments}

We evaluated our proposed approach through both simulation-based experiments and offline map building using real\-world data, both conducted under conditions that included various potential hazards.

In the simulation experiments, we designed a Gazebo environment incorporating static danger sign, dynamic door, and a chair with a seated person to quantitatively compare our method against baseline navigation strategies. The offline map building using real-world data was performed based on data collected in both indoor and outdoor environments, where testing was conducted under conditions that included various potential hazards.

\begin{figure}[t]
    \centering
    \begin{minipage}{0.48\textwidth}  
        \centering
        \includegraphics[width=\textwidth]{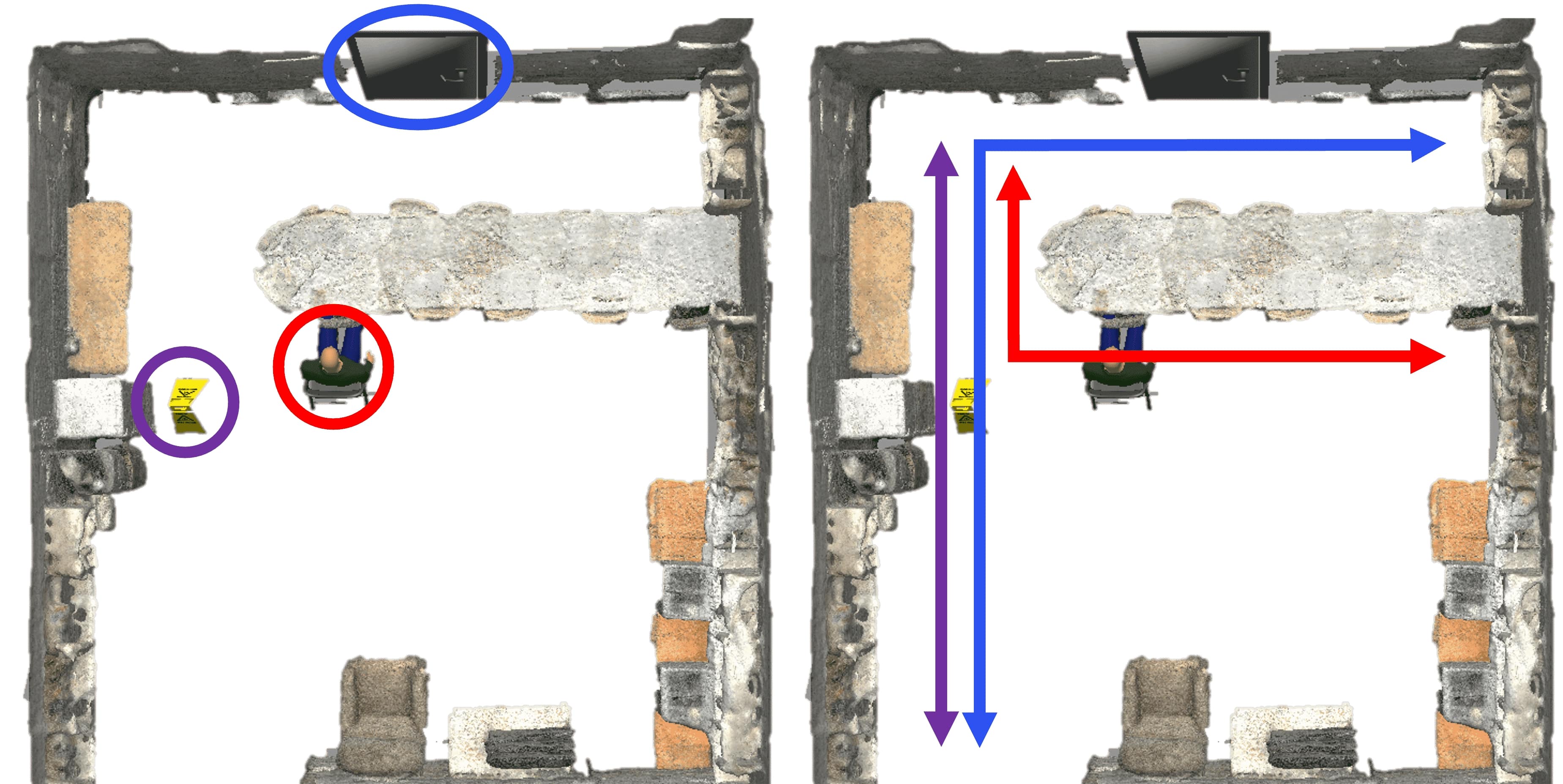}
        \caption{Gazebo simulation environment and corresponding robot trajectories for each scenario. 
        The \textcolor{customPurple}{purple line} represents the robot's trajectory in the \textit{danger sign} scenario, 
        the \textcolor{blue}{blue line} corresponds to the \textit{dynamic door} scenario, 
        and the \textcolor{red}{red line} indicates the path taken in the \textit{seated chair} scenario.}
        \label{fig:experiment_environment}
    \end{minipage}
    \hfill
    \vspace{0.5cm} 
    
    \begin{minipage}{0.45\textwidth}  
        \centering
        \begin{subfigure}[b]{0.32\textwidth} 
            \centering
            \includegraphics[width=\textwidth]{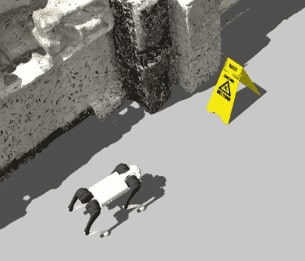}
            \caption{danger sign}
            \label{fig:danger_sign}
        \end{subfigure}
        % \hspace{0.01\textwidth}  
        \begin{subfigure}[b]{0.32\textwidth} 
            \centering
            \includegraphics[width=\textwidth]{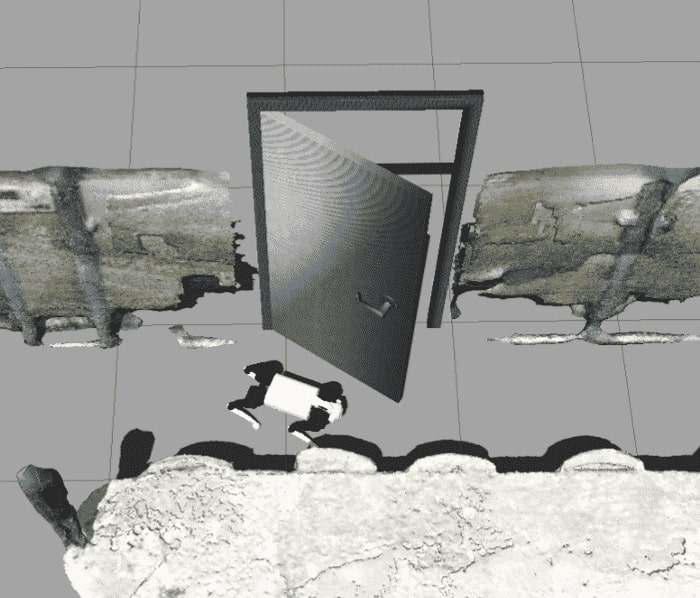}
            \caption{dynamic door}
            \label{fig:dynamic_door}
        \end{subfigure}
        % \hspace{0.01\textwidth} 
        \begin{subfigure}[b]{0.32\textwidth} 
            \centering
            \includegraphics[width=\textwidth]{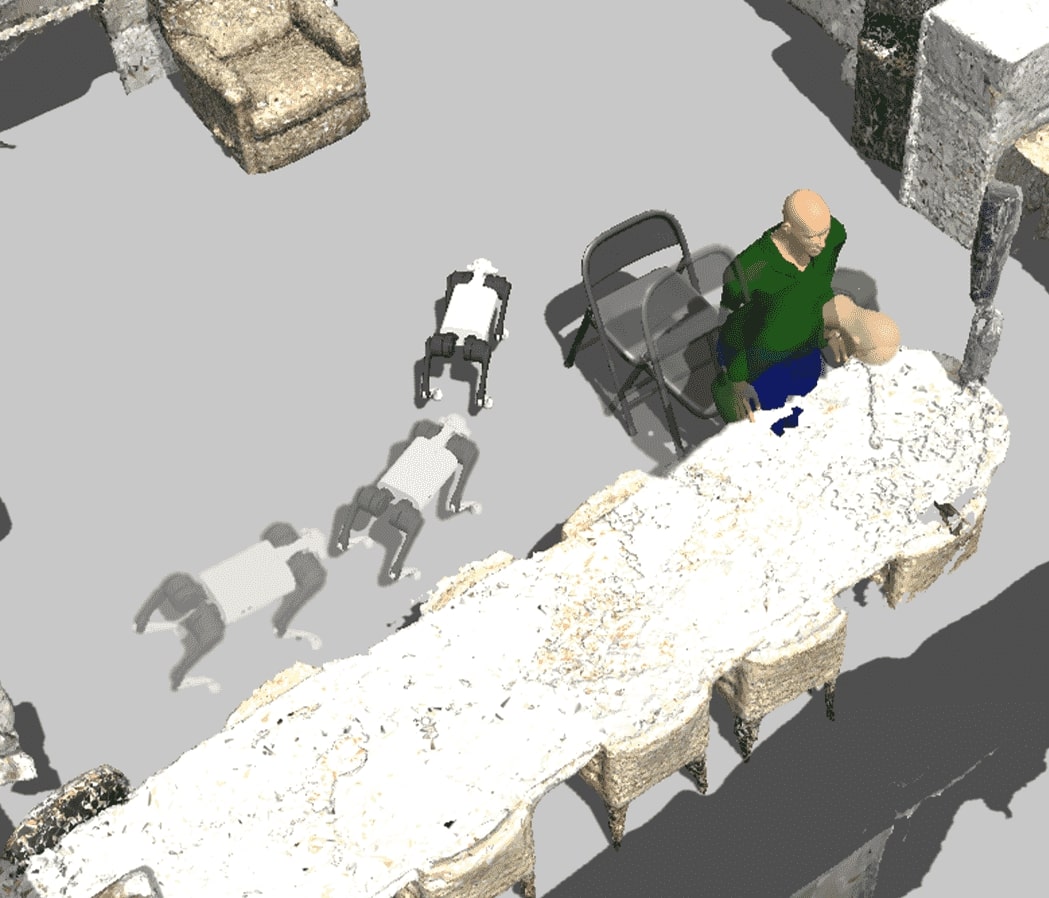}
            \caption{seated chair}
            \label{fig:seated_chair}
        \end{subfigure}
    \end{minipage}
    
    \caption{Three experimental scenarios.}
    \label{fig:three_images}
\vspace{-1.0em}
    
\end{figure}

\subsection{Simulation Experiments}

We used the Gazebo simulator to replicate a room based on real-world measurements. The robot platform was a quadruped robot \textit{Unitree Go1}, which executed the \textit{D* Lite} algorithm \cite{stentz1994d} for path planning and \textit{MPPI} for control \cite{williams2016aggressive}. Localization data was obtained directly from the simulator. Three representative scenarios were selected to evaluate our method’s capacity to anticipate and avoid hazards.

We incorporated three different potential hazards to simulate everyday occurrences: (1) a newly appearing \textit{Danger Sign}, assumed to appear suddenly after the initial map was built, (2) a \textit{Dynamic Door} that opens abruptly as the robot gets close, and (3) a \textit{Seated Chair} with a person who might suddenly stand up, causing the chair to move backward when the robot approaches. Each scenario was tested by guiding the robot along a designated route in both forward and reverse directions, completing ten runs per scenario. The detailed robot trajectories for each scenario are shown in Fig. 4.

Two baseline methods were compared against our approach: a \textit{Geometric Map}, which continuously updates an occupancy grid from depth images, and \textit{E2Map}~\cite{kim2024e2map}, which relies on a pre-built map and updates map costs only upon experiencing specific events such as collisions. We measured the \textit{success rate}, defined as the number of successful runs out of ten in which the robot reached the goal without collisions.

In the first scenario, where a previously nonexistent danger sign suddenly appeared, both the geometric map and our method successfully avoided collisions by refining the map in real time. However, E2Map failed once because it does not update costs until a detrimental event such as collision has already occurred. Next, we tested two scenarios involving hidden or dynamic risks: the seated chair and the dynamic door. Our approach identified each as hazardous without needing prior collision. In particular, it generated hazard reasoning such as \textit{“A person might stand up suddenly from the chair”} and \textit{“A closed door may open abruptly,”} assigning costs to these regions in advance so the robot could circumvent them entirely. In contrast, the geometric map consistently failed because it lacked semantic understanding, while E2Map managed to succeed in 9 out of 10 attempts, requiring at least one event to update the map for future runs. Overall, our method achieved the highest success rate, effectively avoiding dynamic, unseen hazards by leveraging zero-shot hazard inference.

\renewcommand{\arraystretch}{1.3}
\begin{table}[t]
\centering
\caption{Quantitative Results in Simulated Environment.}
\resizebox{1.0\linewidth}{!}{
\begin{tabular}{c|ccc}
Methods &      
    \makecell{danger sign} & 
    \makecell{dynamic door} &
    \makecell{seated chair}
    \\ \hline \hline
    \makecell{Geometric map} & 10/10 & 0/10 & 0/10 \\ \hline
    \makecell{E2Map~\cite{kim2024e2map}} & 9/10 & 9/10 & 9/10 \\ \hline
    \specialrule{0.1em}{0em}{0em}
    \textbf{Ours} & 10/10 & 10/10 & 10/10 \\ \hline
\end{tabular}
}
\label{table:sim_results}
% \vspace{-1.0em}
\end{table}

\begin{figure}[t]
\centering
\includegraphics[width=\linewidth]{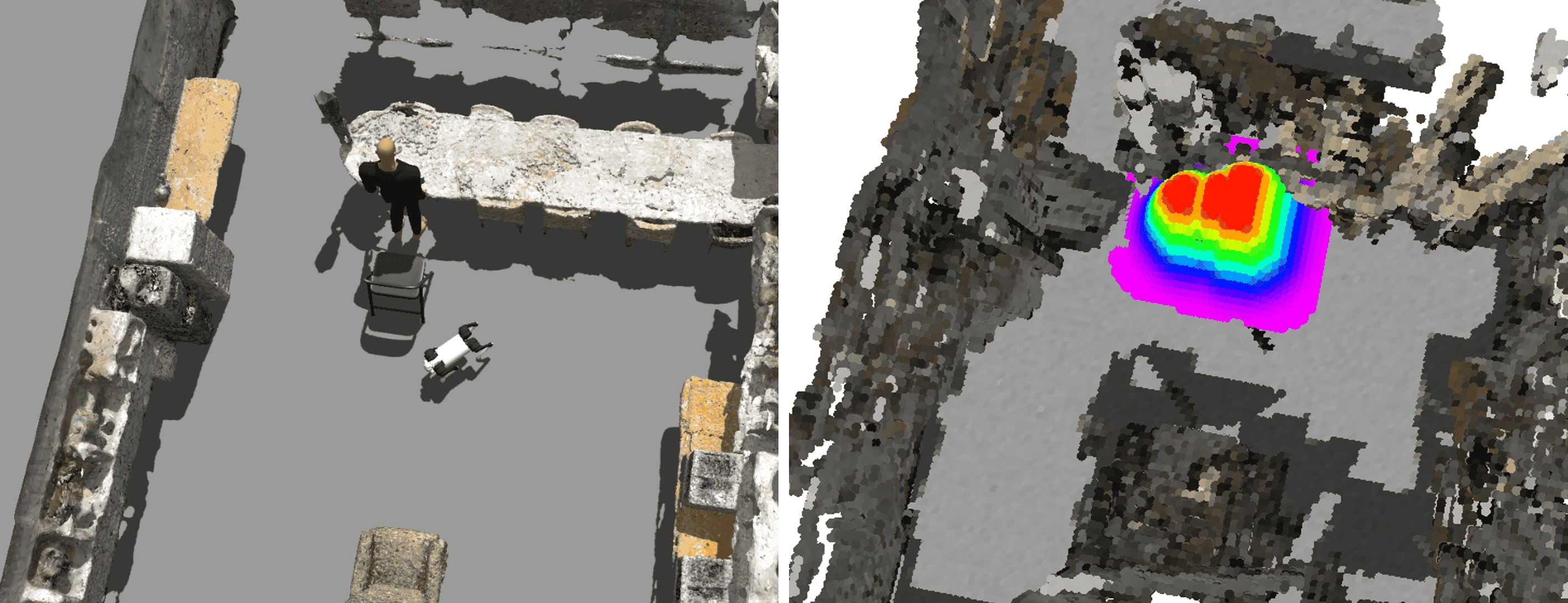}
\caption{\label{fig:qualitative}Qualitative result of the \textit{seated chair} scenario. The robot inferred the chair with a seated person as a potential hazardous object and added costs to the surrounding area, allowing it to navigate safely without collisions.}
\vspace{-1.0em}

\end{figure}

\subsection{Real-World Offline Data Map Building}
For offline data acquisition, we used the \textit{Earth Rover Zero} platform, which was equipped with a ZED stereo camera for RGB-D perception and utilized VINS-Mono \cite{qin2018vins} for pose estimation. Data was recorded using ROS1 \cite{quigley2009ros}. Using the proposed methodology, we generated maps in real-time while incorporating potential hazards in real-world environments, successfully verifying that the system could accurately detect and reflect environmental risks in the mapping process.

\begin{figure}
\centering
\includegraphics[width=\columnwidth]{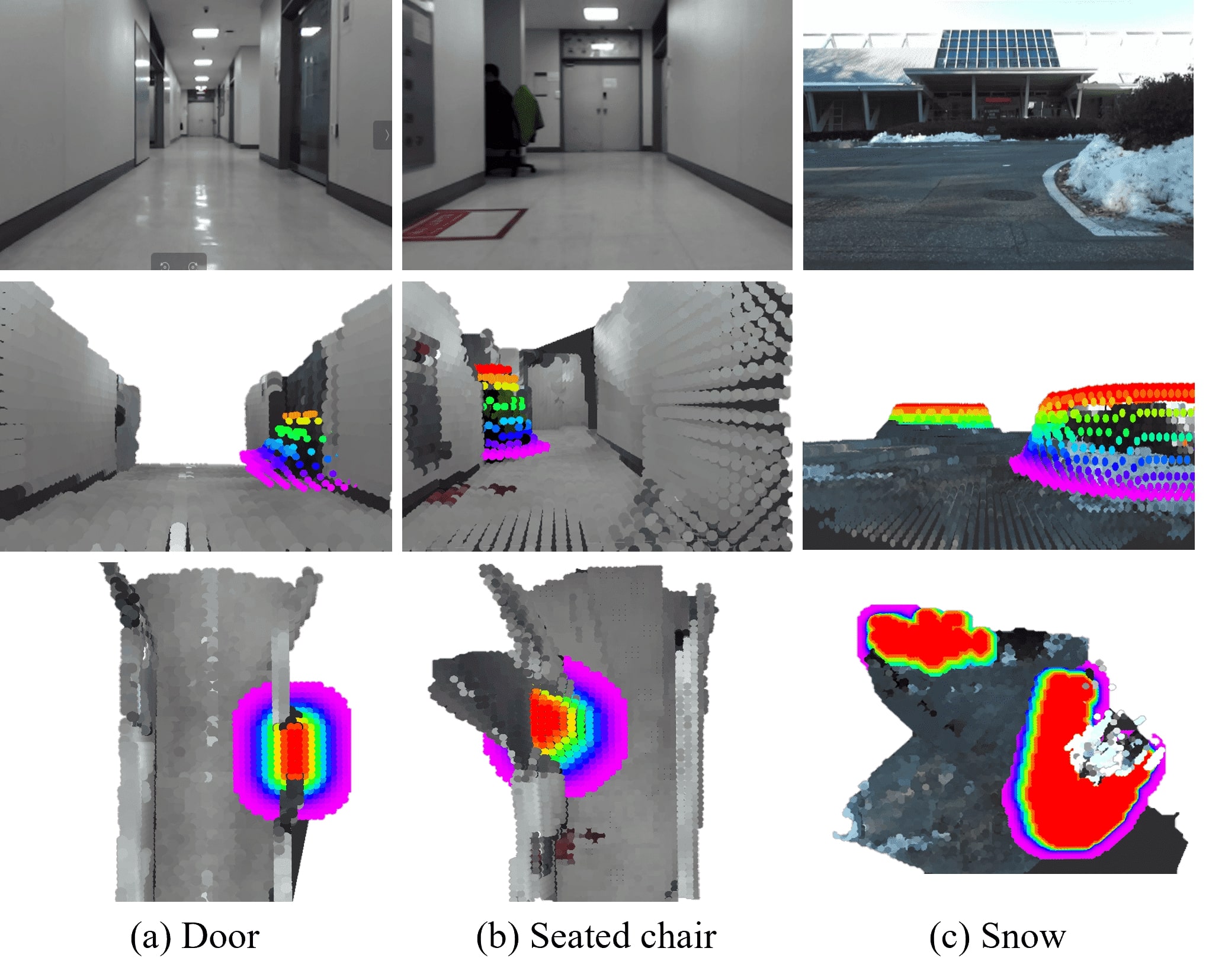}
\caption{\label{fig:ex}Real-World Offline Data Map Building. 
The first row displays the RGB images from the robot’s perspective. The second row visualizes the cost map corresponding to the same viewpoint as the first row. The third row presents the bird’s-eye view (BEV) of the Gaussian map being generated in real-time.
(a) represents the cost visualization of the potential hazard associated with a door in the indoor environment.
(b) shows the cost representation of the potential hazard when a person is seated on a chair in the indoor environment.
(c) illustrates the cost representation of the potential hazard on a snowy surface.}
\vspace{-1.0em}

\end{figure}

In the indoor evaluations, the robot traversed a prototypical hallway populated with multiple latent hazards. The system correctly classified a seated chair as a risk factor—not solely because a chair obstructs the corridor, but because the occupant might rise abruptly, displacing the chair into the robot’s path. Likewise, it flagged a closed door as hazardous by inferring that it could swing open without warning, and it assigned elevated risk to the vicinity of an elevator where pedestrians might emerge suddenly. The resulting cost overlays are visualized in Fig. \ref{fig:bev_map}, demonstrating proactive inflation of navigation costs around each inferred threat.

In the outdoor evaluations, the platform operated on snow-covered terrain. Here, the perception pipeline recognized the snowy patch as a potential slip surface and raised the corresponding traversal cost, prompting the planner to reroute around it. The integrated bird’s-eye-view map in Fig. \ref{fig:ex} confirms that the hazard-aware cost propagation effectively shielded the robot from high-slip regions while preserving global navigability.

By incorporating zero-shot hazard inference and anxiety-based cost mapping, the system demonstrated the ability to enhance safety even in previously unseen hazardous situations. This adaptability suggests that the proposed approach can generalize beyond pre-defined risks, effectively supporting safer robot navigation in dynamic and unpredictable environments. Note that there is a trade-off between safety and efficiency. Using this method may result in slightly longer travel times when no hazards are present. The proposed approach can be considered a priori, whereas E2Map corresponds to a posteriori. By integrating these approaches, safer navigation can be achieved in complex dynamic environments.

\section{Conclusion} 
This work presented a zero-shot, language-driven approach to hazard-aware robot navigation. By integrating VLMs for hazard detection and an emotion evaluator that assigns anxiety-based scores, we proactively estimated risk before direct encounters. Through zero-shot segmentation, the system converted textual cues into spatial cost values, which were then fused with a geometric map to guide navigation. Using the inference capabilities of the VLMs, we successfully detected hazardous situations and assigned corresponding costs on the map, allowing the robot to avoid them. Furthermore, language served as a bridge between the VLM and the zero-shot segmentation process, enabling the system to operate seamlessly under a variety of conditions. Experiments in both real-world and simulated settings demonstrated that our method outperforms baseline approaches, providing more robust and proactive hazard predictions for safer navigation.

\section{ACKNOWLEDGMENT}

This research was supported by the Korean Ministry of Land, Infrastructure and Transport (MOLIT) as the Innovative Talent Education Program for Smart City and Korea Institute for Advancement of Technology (KIAT) grant funded by the Korea Government (MOTIE) (P0017304, Human Resource Development Program for Industrial Innovation). The Institute of Engineering Research at Seoul National University provided research facilities for this work.
This work was supported by the Challengeable Future Defense Technology Research and Development Program
Through the Agency For Defense Development (ADD) funded by the Defense Acquisition Program Administration (DAPA) in 2025 under Grant 915108201.

\bibliographystyle{IEEEtran}
\bibliography{reference}
\end{document}